\documentclass[10pt,twocolumn,letterpaper]{article}

\usepackage{cvpr}
\usepackage{times}
\usepackage{epsfig}
\usepackage{graphicx}
\usepackage{amsmath}
\usepackage{amssymb}
\usepackage{bm}
\usepackage{enumerate}
\usepackage{enumitem}
\DeclareMathOperator*{\argmax}{arg\,max}

\usepackage{color}
\definecolor{mypink}{RGB}{255,135,180}
\definecolor{myyellow}{RGB}{255,200,0}
\definecolor{mygreen}{RGB}{0,100,0}
\definecolor{mygold}{RGB}{255,185,0}
\definecolor{mymaroon}{RGB}{128,0,0}
\definecolor{myaquamarine}{RGB}{120,200,190}
\definecolor{myturquoise}{RGB}{64,224,208}
\definecolor{mymagenta}{RGB}{255,0,255}
\definecolor{myviolet}{RGB}{238,130,238}

\usepackage[pagebackref=true,breaklinks=true,letterpaper=true,colorlinks,bookmarks=false]{hyperref}

\cvprfinalcopy 

\def\cvprPaperID{5585} 

\ifcvprfinal\fi
\begin{document}

\title{Set-Constrained Viterbi for Set-Supervised Action Segmentation}

\author{Jun Li\\
Oregon State University\\
{\tt\small liju2@oregonstate.edu}
\and
Sinisa Todorovic\\
Oregon State University\\
{\tt\small sinisa@oregonstate.edu}
}

\maketitle

\begin{abstract}
   This paper is about weakly supervised action segmentation, where the ground truth specifies only a set of actions present in a training video, but not their true temporal ordering.  Prior work typically uses a classifier that independently labels video frames for generating the pseudo ground truth, and multiple instance learning for training the classifier. We extend this framework by specifying an HMM, which accounts for co-occurrences of action classes and their temporal lengths, and by explicitly training the HMM on a Viterbi-based loss. Our first contribution is the formulation of a new set-constrained Viterbi algorithm (SCV). Given a video, the SCV generates the MAP action segmentation that satisfies the ground truth. This prediction is used as a framewise pseudo ground truth in our HMM training. Our second contribution in training is a new regularization of feature affinities between training videos that share the same action classes. Evaluation on action segmentation and alignment on the Breakfast, MPII Cooking2, Hollywood Extended datasets demonstrates our significant performance improvement for the two tasks over prior work.
\end{abstract}


\section{Introduction}
This paper addresses action segmentation by labeling video frames with action classes under set-level weak supervision in training. Set-supervised training means that the ground truth specifies only a set of actions present in a training video. Their temporal ordering and the number of their occurrences remain unknown. This is an important problem arising from the proliferation of big video datasets where providing detailed annotations of a temporal ordering of actions is prohibitively expensive. One example application is action segmentation of videos that have been retrieved from a dataset based on word captions \cite{dong2019dual, Shao_2018_ECCV}, where the captions do not describe temporal relationships of actions.


There is scant work on this problem. Related work \cite{richard2018action} combines a frame-wise classifier with a Hidden Markov Model (HMM). The classifier assigns action scores to all frames, and the HMM models a grammar and temporal lengths of actions. However, their HMM and classifier are not jointly learned, and performance is significantly below that of counterpart approaches with access to ground-truth temporal ordering of actions in training \cite{Li_2019_ICCV}.

In this paper, we also adopt an HMM model that is grounded on a  fully-connected two-layer neural network which extracts frame features and scores label assignments to frames. Our HMM models temporal extents and co-occurrence of actions. We jointly train the HMM and neural network on a maximum posterior probability (MAP) action assignment to frames that satisfy the set-level ground truth of a training video. We expect that the inferred MAP action sequence is more optimal for joint training than the pseudo ground truth generated independently for each frame as in \cite{richard2018action}.  We cast this MAP inference as a set-constrained structured prediction of action labels such that every label from the ground-truth action set appears at least once in the MAP prediction. This problem has been shown to be NP-hard \cite{DBLP:journals/corr/BilgeCGSAE15}. Therefore, our main novelty is an efficient approximation of this NP-hard problem for our set-supervised training.

\begin{figure}
\begin{center}
\includegraphics[scale=0.4]{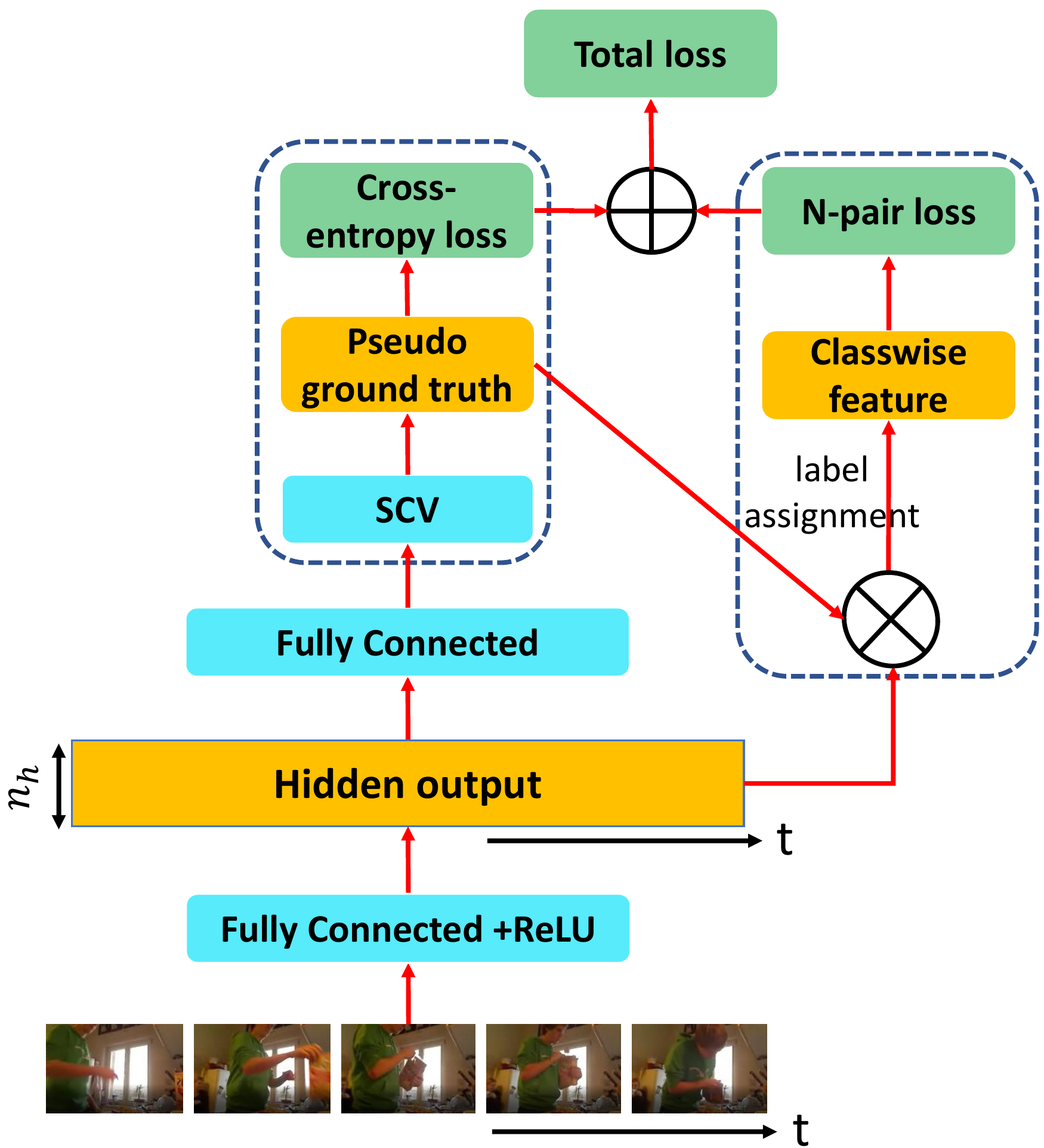}
\end{center}
   \caption{Our two contributions for set-supervised training include: (1) Set Constrained Viterbi (SCV) algorithm for estimating the MAP pseudo ground truth, and (2) Regularization for maximizing a margin between shared action classes and non-shared action classes in training videos. The pseudo ground truth is used for computing the cross-entropy loss and n-pair loss.}
\label{fig:Overview}
\end{figure}

Fig.~\ref{fig:Overview} illustrates our two  contributions. First, we propose an efficient Set Constrained Viterbi (SCV) algorithm for the MAP inference on training videos. Our SCV consists of two steps. In the first step, we run the Viterbi algorithm for inferring a MAP action assignment to frames, where the predicted actions are restricted to appear in the ground-truth set.  When our initial prediction is missing some actions from the ground truth, we proceed to our second step that sequentially flips the predicted labels of lowest scoring video segments into the missing labels, so as to minimally decrease the assignment's posterior probability. This is done until the predicted sequence contains all actions from the ground truth. While this second step could alternatively use a more principled algorithm for enforcing the ground-truth constraint (e.g., Markov Chain Monte Carlo sampling), such an alternative would be significantly more complex and hence poorly justified.

Our second contribution is that we specify a new regularization of our set-supervised learning.
This regularization is aimed at our two-layer neural network for extracting frame features. The regularization enforces that frame features are closer for frames belonging to the same actions than other frames of different actions. Our key novelty is a new n-pair loss that deals with sequences of actions, unlike the standard n-pair loss used for distance-metric learning of images or videos with a single class label. For pairs of training videos with generally different ground-truth sets of actions, our n-pair loss minimizes and maximizes distances of frame features over shared and non-shared actions, respectively.


For evaluation, we address action segmentation and action alignment  on the Breakfast, MPII Cooking2, Hollywood Extended  datasets. As in \cite{richard2018action}, action alignment means that for a test video we know the set of actions present, and the task is to accurately label frames with these actions. Note that this definition increases difficulty over the definition of action alignment used in prior work \cite{richard2017weakly,richard2018neuralnetwork}, where for a test video we would also know the temporal ordering of actions present. Our experiments demonstrate that we significantly outperform existing work for the two tasks.

The rest of the paper is organized as follows.  Sec.~\ref{sec:Related Work} reviews related work,  Sec.~\ref{sec:Problem Setup} formalizes our problem, Sec.~\ref{sec:Methods} specifies our SCV and regularization, Sec.~\ref{sec:inference} describes our inference in testing, and Sec.~\ref{sec:Experiments} presents our experiments.

\section{Related Work}\label{sec:Related Work}
This section reviews closely related work on weakly supervised: (a) Action segmentation and localization; (b) Distance metric learning over sequences of labels. A review of both fully supervised and unsupervised action segmentation 
is beyond our scope, for obvious reasons.

\textbf{Learning with the Temporal Ordering Constraint.} A number of related approaches assume access to the temporal ordering of actions in training \cite{bojanowski2014weakly, huang2016connectionist, koller2016deep, koller2017re, kuehne2017weakly, richard2017weakly, ding2018weakly, richard2018neuralnetwork, chang2019d}. For example,  Huang \etal \cite{huang2016connectionist} propose an extended connectionist temporal classification for taking into account temporal coherence of features across consecutive frames. Bojanowski \etal \cite{bojanowski2014weakly} formulate a convex relaxation of discriminative clustering based on a conditional gradient (Frank-Wolfe) algorithm for action alignment. Other approaches \cite{richard2017weakly,richard2018neuralnetwork} use a statistical language model for action segmentation. All these methods in training generate a framewise pseudo ground truth which satisfies the provided ground-truth action ordering. As our training is inherently less constrained, for generating a pseudo ground truth, we face a much larger search space of label sequences that satisfy our ground truth. This informs our design choices to adopt simpler models than theirs (e.g., our two-layer network vs. a more complex recurrent network in \cite{richard2018neuralnetwork}) so as to facilitate our  set-supervised learning.

\textbf{Set-supervised Learning.} Action localization under unordered set-level supervision has recently made significant progress \cite{sun2015temporal, singh2017hide, wang2017untrimmednets, shou2018autoloc, paul2018w,jhuang2011large, soomro2012ucf101}. All of these methods, however, address videos with few action classes and a lot of background frames. In contrast, we consider a more challenging setting, where videos show significantly more action classes. As shown in \cite{richard2018action}, these methods are particularly designed to detect sparse actions in relatively long background intervals, and thus are not suitable for our setting of densely labeled videos. For example, Shou \etal \cite{shou2018autoloc} use the outer-inner-contrastive loss for detecting action boundaries, and Paul \etal \cite{paul2018w} minimize an action affinity loss for multi-instance learning (MIL) \cite{zhou2002neural}. The affinity loss in \cite{paul2018w} is related to our regularization. However, they enforce only similarity between video features of the same action class, whereas our more general n-pair loss additionally maximizes a distance between features of frames belonging to non-shared action classes.

The most related approach \cite{richard2018action} uses MIL \cite{zhou2002neural} to train a frame-wise classifier. 
Instead, we formulate the new set-constrained Viterbi algorithm to predict MAP action sequences for the HMM training, as well as regularize our training with distance-metric learning on frame features.

\section{Our Problem Setup and Model}\label{sec:Problem Setup}
\textbf{Problem Formulation.} A training video of length $T$ is represented by a sequence of frame features, $\bm{x} = [x_1,...,x_t,...,x_T]$, and annotated with an unordered set of action classes $C=\{c_1,\cdots,c_m,\cdots, c_M\}$, where $C$ is a subset of a large set of all actions, $C\subseteq\mathcal{C}$. $T$ and $M$ may vary across the training set. There may be multiple instances of the same action in a training video.
For a given video, our goal is to find an optimal action segmentation, $(\hat{\bm{c}}, \hat{\bm{l}})$, where $\hat{\bm{c}}=[\hat{c}_1,...,\hat{c}_n,...,\hat{c}_{\hat{N}}]$ denotes the predicted temporal ordering of action labels of length $\hat{N}$, $\hat{c}_n\in\mathcal{C}$, and $\hat{\bm{l}} = [\hat{l}_1,\cdots,\hat{l}_{\hat{N}}]$ are their corresponding temporal extents.

\textbf{Our Model.} We use an HMM to estimate the MAP  $(\hat{\bm{c}}, \hat{\bm{l}})$ for a given video $\bm{x}$ as
\begin{equation}
\arraycolsep=1pt
\begin{array}{lcl}
\label{eq:Markov}
(\hat{\bm{c}}, \hat{\bm{l}}) &=&  \displaystyle\argmax_{N, \bm{c},\bm{l}}\; p(\bm{c},\bm{l}|\bm{x})
=\displaystyle\argmax_{N,\bm{c},\bm{l}}\; p(\bm{c})p(\bm{l}|\bm{c})p(\bm{x}|\bm{c},\bm{l}), \\
&= &  \displaystyle\argmax_{N,\bm{c},\bm{l}}\Big[\prod_{n=1}^{N-1} p({c_{n+1}}|c_n)\Big] \Big[\prod_{n=1}^{N}p(l_n|c_n)\Big] \\
  & & \quad\quad \quad\quad \displaystyle \cdot \Big[\prod_{t=1}^{T}p(x_t|c_{n(t)})\Big].
 \end{array}
\end{equation}
A similar HMM formulation is used in \cite{richard2018action}, but with a notable difference in the definition of $p(\bm{c})$. In \cite{richard2018action},  $p(\bm{c})$ is equal to a positive constant for all legal action grammars, where they declare $\bm{c}$ as a legal grammar if all actions in $\bm{c}$ regardless of their ordering are seen in at least one ground truth of the training dataset. In contrast, we specify $p(\bm{c})$ in \eqref{eq:Markov} as a product of transition probabilities between actions along the sequence $\bm{c}$, $\prod_{n=1}^{N-1} p({c_{n+1}}|c_n)$. Another difference from \cite{richard2018action} is that we consider both {\em static} and {\em dynamic} formulations of the HMM, and \cite{richard2018action} uses only the former.

 In this paper, we specify and evaluate two distinct versions of the HMM given by \eqref{eq:Markov} -- namely, the static and dynamic HMM. The static probabilities in \eqref{eq:Markov} are pre-computed directly from the available set-supervised training ground truth. The dynamic probabilities in \eqref{eq:Markov}  are iteratively updated based on the previous MAP assignments $\{(\hat{\bm{c}}, \hat{\bm{l}})\}$ over all training videos.

\textbf{Static HMM.} The static transition probability $p^s({c_{n+1}}|c_n)$ in \eqref{eq:Markov} is defined as
\begin{align}\label{eq:transition static}
    p^s({c_{n+1}}|c_n) = \#({c_{n+1}}, c_n) ~/ ~\#(c_n),
\end{align}
where $\#(\cdot)$ denotes the number of action or action-pair occurrences in the training ground truth.

The static likelihood of action lengths $p^s(l|c)$ in \eqref{eq:Markov} is modeled as a class-specific Poisson distribution:
\begin{align}
    p^s(l|c) = \frac{(\lambda_{c}^s)^{l}}{l!}e^{-\lambda_{c}^s},
    \label{eq:Poisson}
\end{align}
where $\lambda_{c}^s$ is the static, expected temporal length of  $c \in \mathcal{C}$. For estimating $\lambda_{c}^s$, we ensure that, for all training videos indexed by  $v$, the accumulated mean length of all classes from the ground truth $C_v$ is close to the video's length $T_v$. This is formalized as minimizing the following quadratic objective:
\begin{align}\label{eq:fixed length}
  \text{minimize} \quad  \sum_{v}(\sum_{c\in C_v} \lambda_c^s - T_v)^2, \quad \text{s.t.}\quad  \lambda_c^s > l_{\text{min}},
\end{align}
where $l_{\text{min}}$ is the minimum allowed action length which ensures a reasonable action duration.

The static likelihood $p^s(x|c)$ in  \eqref{eq:Markov} is estimated as:
\begin{align}
    p^s(x|c) \propto \frac{p(c|x)}{p^s(c)},\quad\quad p^s(c) = \frac{\sum_{v}T_v\cdot 1(c \in C_v)}{ \sum_{v}{T_{v}}}.
    \label{eq:likelihood_static}
\end{align}
where $p(c|x)$ is the softmax score of our neural network, and  $1(\cdot)$ is the indicator function. The static class prior per frame $p^s(c)$ in \eqref{eq:likelihood_static} is a percentage of the total footage of training videos having $c$ in their ground truth.

 \textbf{Dynamic HMM.} The dynamic transition probability $p^d({c_{n+1}}|c_n)$ in \eqref{eq:Markov} is defined as
\begin{align}\label{eq:transition dynamic}
    p^d(c_{n+1} = \hat{c}_{n+1}|c_n =\hat{c}_{n} ) = \#({\hat{c}_{n+1}}, \hat{c}_n) ~/ ~\#(\hat{c}_n),
\end{align}
where $\#({\hat{c}_{n+1}}, \hat{c}_n)$ and $\#(\hat{c}_n)$ are the numbers of predicted consecutive pairs and single actions, respectively, in
the previous MAP assignments $\{(\hat{\bm{c}}, \hat{\bm{l}})\}$ over all training videos.

The dynamic likelihood of action lengths in \eqref{eq:Markov} is also the Poisson distribution, $p^d(l|c)=\frac{(\lambda_{c}^d)^{l}}{l!}e^{-\lambda_{c}^d}$, where for the previous MAP assignments $\{(\hat{\bm{c}}, \hat{\bm{l}})\}$ over all training videos,  we estimate the expected temporal length of $c \in \mathcal{C}$ as
\begin{equation}
    \lambda_c^d =\displaystyle \Big[\sum_{v}\sum_{n=1}^{\hat{N}} \hat{l}_{v,n}\cdot 1(c=\hat{c}_{v,n})\Big] / \sum_{v}\sum_{n=1}^{\hat{N}} 1(c=\hat{c}_{v,n}).
    \label{eq:dynamic length}
\end{equation}

Finally, $p^d(x|c)$ in  \eqref{eq:Markov} is defined as
\begin{align}
    p^d(x|c) \propto \frac{p(c|x)}{p^d(c)},\quad p^d(c) = \frac{\sum_{v}\sum_{n=1}^{\hat{N}} \hat{l}_{v,n}\cdot 1(c=\hat{c}_{v,n})}{ \sum_{v}{T_{v}}}.
    \label{eq:likelihood_dynamic}
\end{align}
where $p^d(c)$ is a percentage of training frames predicted to belong to class $c$ in the previous MAP assignments $\{(\hat{\bm{c}}, \hat{\bm{l}})\}$ over all training videos.


\textbf{Two-layer Neural Network.} Our HMM is grounded on the frame features  $\bm{x}$ via a two-layer fully connected neural network. The first fully connected layer uses ReLU to extract hidden features $\bm{h}$ as
\begin{align}
    \bm{h} = \max (0, \bm{W}^1\bm{x} \oplus \bm{b}^1),
    \label{eq:hidden}
\end{align}
where $\bm{x} \in\mathbb{R}^{d \times T}$ denotes $d$-dimensional unsupervised features of $T$ frames, $\bm{W}^1 \in \mathbb{R}^{n_h \times d}$ for $n_h=256$ hidden units, $\bm{b}^1 \in \mathbb{R}^{n_h \times 1}$, $\bm{h} \in \mathbb{R}^{n_h\times T}$, and $\oplus$ indicates that the bias $\bm{b}^1$ is added to every column in the matrix $\bm{W}^1\bm{x}$.

The second fully connected layer computes the matrix of unnormalized scores for each action class in $\mathcal{C}$ as
\begin{align}
    \bm{f} = \bm{W}^2\bm{h} \oplus \bm{b}^2,
    \label{eq:secondlayer}
\end{align}
where $\bm{W}^2 \in \mathbb{R}^{|\mathcal{C}|\times n_h}$, $\bm{b}^2 \in \mathbb{R}^{|\mathcal{C}| \times 1}$,  and $\bm{f} \in \mathbb{R}^{|\mathcal{C}|\times T}$. Thus, an element  $\bm{f}[c,t]$ of matrix $\bm{f}$ represents the network's unnormalized score for class $c$ at video frame $t$.

We use both $\bm{h}$ and $\bm{f}$ for estimating loss in our training.

\section{Our Set-Supervised Training}\label{sec:Methods}
For our training, we generate the MAP labels $(\hat{\bm{c}}, \hat{\bm{l}})$ for every training video by using our new Set-Constrained Viterbi algorithm (SCV). $(\hat{\bm{c}}, \hat{\bm{l}})$ is taken as the pseudo ground truth and used to estimate the cross-entropy loss and n-pair regularization loss for updating parameters of the HMM and two-layer neural network. In the following, we first formulate our SCV, and then the two loss functions.

\subsection{Set Constrained Viterbi}
\begin{figure}
\begin{center}
\includegraphics[width=0.95\linewidth]{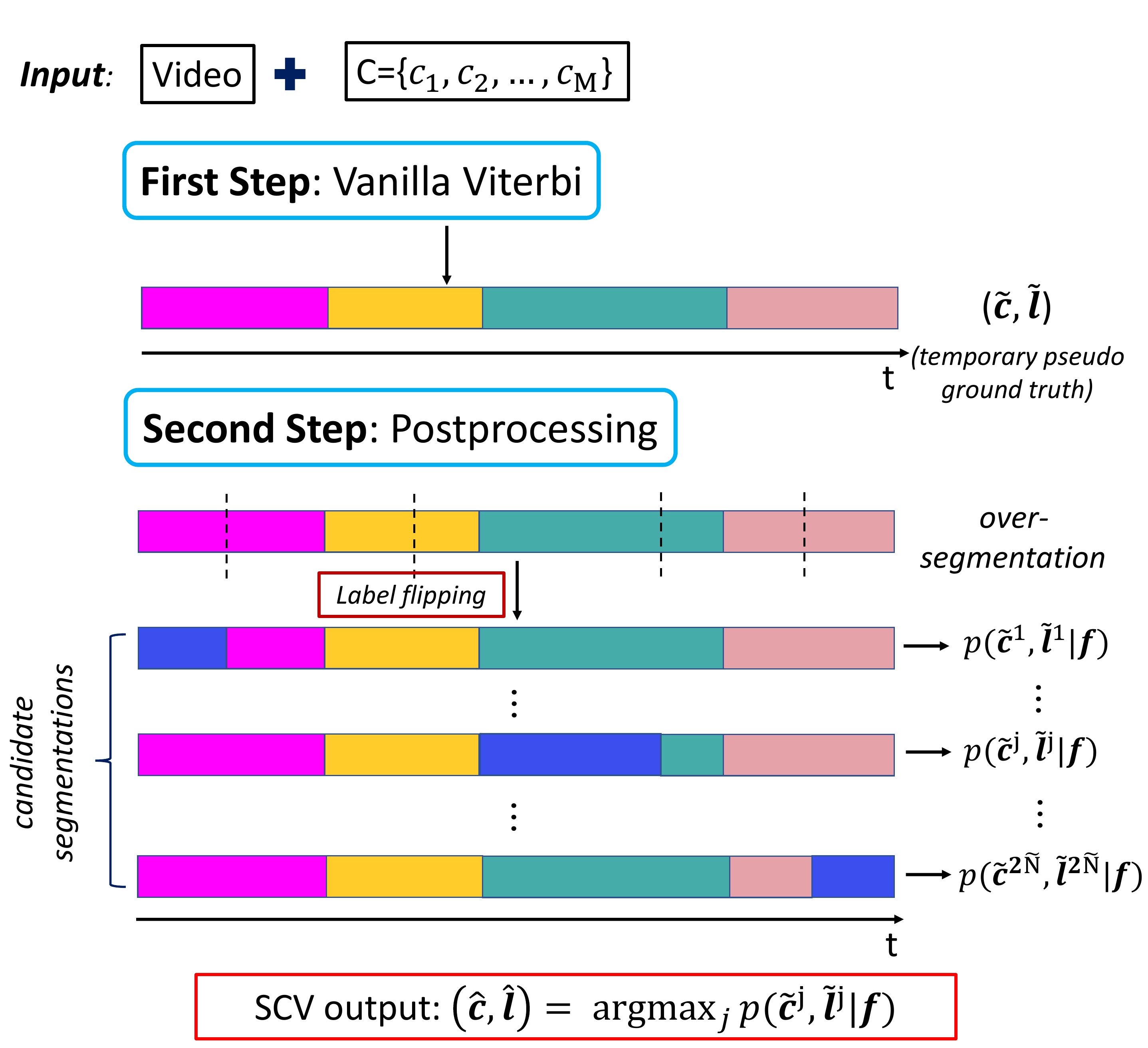}
\end{center}
   \caption{Our SCV consists of two steps shown top-to-bottom. The first step is the vanilla Viterbi algorithm that infers an initial MAP action sequence $(\tilde{\bm{c}}, \tilde{\bm{l}})$. $\tilde{\bm{c}}$ may be missing some classes (e.g., "blue" label) from the ground truth $C$. The second step sequentially flips labels of video oversegments to the missing labels,  such that the MAP score is minimally reduced, until the final prediction $(\hat{\bm{c}}, \hat{\bm{l}})$ includes all actions from $C$.}
\label{fig:SCV}
\end{figure}

Inference in training is formulated as the NP-hard all-color shortest path problem \cite{DBLP:journals/corr/BilgeCGSAE15}.  Given a training video and its ground truth $C$, our goal is to predict the MAP $(\hat{\bm{c}}, \hat{\bm{l}})$, such that every ground-truth action $c\in C$ occurs at least once in $\hat{\bm{c}}$. As shown in Fig.~\ref{fig:SCV}, our efficient solution to this NP-hard problem consists of two steps.

{\bf In the first step}, we ignore the all-color constraint, and conduct the HMM inference by using the vanilla Viterbi algorithm for predicting an initial MAP $(\tilde{\bm{c}}, \tilde{\bm{l}})$, as specified in \eqref{eq:Markov}, where actions in $\tilde{\bm{c}}$ are constrained to come from $C$. From \eqref{eq:Markov} and \eqref{eq:secondlayer},  our first step is formalized as
\begin{align}
     (\tilde{\bm{c}}, \tilde{\bm{l}}) = 
           \argmax_{\substack{N,\bm{c},\bm{l}\\ \bm{c}\in C^N}}&\Big[\prod_{n=1}^{N-1} p({c_{n+1}}|c_n)\Big]  \Big[\prod_{n=1}^{N}p(l_n|c_n)\Big] \nonumber \\ & \cdot \Big[\prod_{t=1}^{T}\frac{p(\bm{f}[c_{n(t)},t])}{p(c_{n(t)})}\Big],
           \label{eq:FirstStep}
\end{align}
where $p(\bm{f}[c_{n(t)},t])$ denotes the softmax score for class $c_n\in C$ at frame $t$. For estimating the likelihood of action lengths $p(l_n|c_n)$ in \eqref{eq:FirstStep}, we use the Poisson distribution, parameterized by either the static expected length $\lambda_c^s$ as in \eqref{eq:fixed length} or dynamic $\lambda_c^d$ as in \eqref{eq:dynamic length}. For the transition probabilities in \eqref{eq:FirstStep}, $p({c_{n+1}}|c_n)$, we use either the static definition given by \eqref{eq:transition static} or the dynamic definition given by \eqref{eq:transition dynamic}.

The vanilla Viterbi begins by computing: (i) Softmax scores $p(\bm{f}[c_{n(t)},t])$ with the two-layer neural network for $t=1,\dots,T$, and (ii) Per-frame class priors $p(c_{n(t)})$ for all $c\in\mathcal{C}$ as in \eqref{eq:likelihood_static} for the static HMM or in \eqref{eq:likelihood_dynamic} for the dynamic HMM.
Then, the algorithm uses a recursion to efficiently compute $(\tilde{\bm{c}}, \tilde{\bm{l}})$. Let $\bm{\ell}[c,t]$  denote the maximum log-posterior of all action sequences $\{\bm{c}\}$ ending with action $c \in C$ at video frame $t$, $t=1,\dots,T$. 
From \eqref{eq:FirstStep},  $\bm{\ell}[c,t]$ can be recursively estimated as
\begin{align}
    \bm{\ell}[c,t]  =  \max_{\substack{t'<t\\  c'\neq c \\ c'\in C}} & \Big[\bm{\ell}[c',t'] {+} \log p(t{-}t'|c) {+} \log p(c|c')\nonumber \\
     & {+}\sum_{k = t'+1}^{t} \log \frac{p(\bm{f}[c,k])}{p(c)}\Big],
\end{align}
where $\bm{\ell}[c,0]=0$ and $p(t{-}t'|c)$ is the Poisson likelihood that action $c$ has length $(t{-}t')$. During the recursion, we discard action sequences ${\bm{c}}$, whose accumulated mean length exceeds the video length $T$, \ie  $\sum_{c\in {\bm{c}}}\lambda_c>T$.

After estimating the entire matrix of maximum log-posteriors, $\bm{\ell}[c,t]$, the Viterbi algorithm back-traces the optimal path in the matrix, starting from the maximum element of the last video frame  $\max_{c\in C} \bm{\ell}[c,T]$, resulting in $(\tilde{\bm{c}}, \tilde{\bm{l}})$.


{\bf The second step} is aimed at satisfying the ground-truth constraint by adding labels from $C$ that have been missed in $(\tilde{\bm{c}}, \tilde{\bm{l}})$. If $\tilde{\bm{c}}$ consists of all classes from the ground truth $C$, we stop our inference. Otherwise, we proceed by sequentially searching for the missing classes. We first oversegment the video in an unsupervised manner, as explained below. Then, among many candidate oversegments, the best oversegment is selected for replacing its predicted class $\tilde{c}$ with the best missed class $c'\in C$ and $c'\notin \tilde{\bm{c}}$, such that the resulting posterior $p(\tilde{\bm{c}}', \tilde{\bm{l}}'|\bm{f})$ is minimally decreased relative to $p(\tilde{\bm{c}}, \tilde{\bm{l}}|\bm{f})$. Note that, due to the splitting of video segments into oversegments, our class flipping of oversegments does not remove the predicted classes that belong to $C$. The class flipping is sequentially conducted until all classes from $C$ appear in the solution. The final solution is  $(\hat{\bm{c}}, \hat{\bm{l}})$.

For efficiency, our unsupervised video oversegmentation is a simple splitting of every predicted video segment in $\tilde{\bm{l}}$ into two parts. Specifically, for every $n$th predicted video segment of length $\tilde{l}_n$ starting at time $t_n<T$, we estimate the cosine similarities of consecutive frames as $\{\bm{h}[t_n+k]^\top\bm{h}[t_n+k+1]: k=0,\dots,\tilde{l}_n-2\}$, where $\bm{h}$ is specified in \eqref{eq:hidden}. Then, we select the $\hat{k}$th frame  with the minimum similarity to its neighbor to split the interval into two oversegments $[t_n,t_n+\hat{k}]$ and $[t_n+\hat{k}+1,t_n+\tilde{l}_n - 1]$.

\subsection{The Cross-Entropy Loss}

After the SCV infers the pseudo ground truth of a given training video,  $(\hat{\bm{c}}, \hat{\bm{l}})$, we compute the incurred cross-entropy loss as
\begin{align}\label{eq:cross entropy}
    \mathcal{L}_{\text{CE}} = - \sum_{t=1}^{T}\log p(\hat{c}_{n(t)}|x_t), \;\hat{c}_{n(t)}\in \hat{\bm{c}}
\end{align}
where $p(\hat{c}_{n(t)}|x_t) = p(\bm{f}[\hat{c}_{n(t)},t])$ is the softmax score of the two-layer neural network for the $n$th  predicted class $\hat{c}_{n(t)}\in \hat{\bm{c}}$ at frame $t$.

\subsection{Regularization with the N-Pair Loss}\label{sec:n-pair}
Our training is regularized with the n-pair loss aimed at minimizing a distance between pairs of training videos sharing action classes in their respective set-level ground truths.  We expect that the proposed regularization would help constrain our set-supervised problem in training.

For every pair of training videos $v$ and $v'$, we consider an intersection of their respective ground truths:  $C_{vv'} = C_{v} \cap C_{v'}$. When  $C_{vv'}\ne \emptyset$, for every shared class  $c\in C_{vv'}$, our goal is two-fold:
\begin{enumerate}[itemsep=-1pt,topsep=2pt, partopsep=1pt]
\item Minimize a distance, $d_{vv'}^{cc}$, between features of $v$ and $v'$ that are relevant for predicting the shared class $c\in C_{vv'}$, and simultaneously.
\item Maximize distances $d_{vv'}^{ac}$ and $d_{vv'}^{cb}$ between features of the shared class $c\in C_{vv'}$ and  features of non-shared classes $a\in C_{v} \setminus C_{vv'}$ and $b\in C_{v'} \setminus C_{vv'}$.
\end{enumerate}
%
%
%
\begin{figure}[t]
\begin{center}
\includegraphics[width=0.95\linewidth]{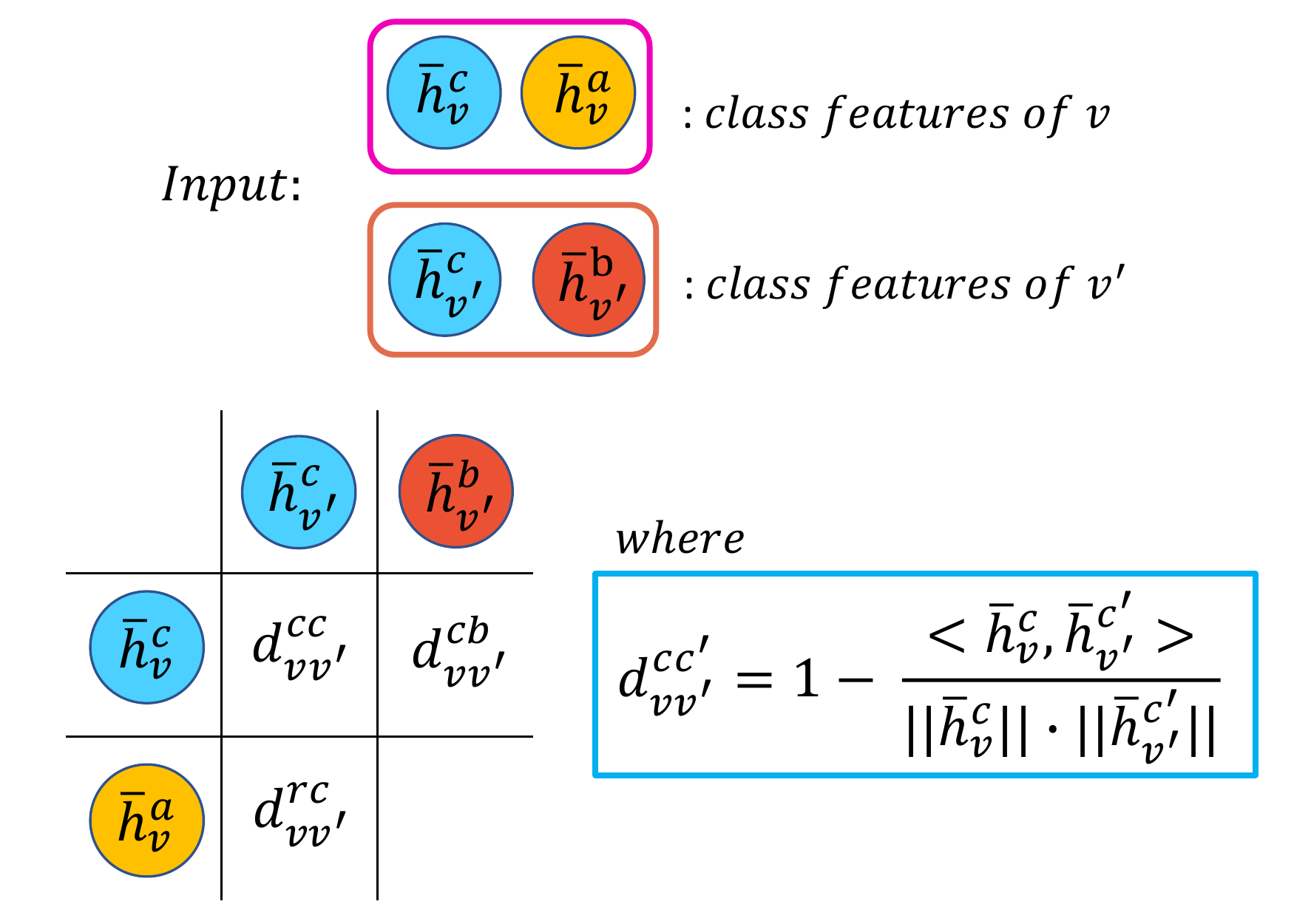}
\end{center}
   \caption{An example of estimating three cosine distances that are used for computing the n-pair loss. Given video $v$ with only two classes $\{c,a\}$ and another video $v'$ with also two classes $\{c,b\}$, we first compute their average class features: $\{\bar{h}_v^c,\bar{h}_v^a\}$ and $\{\bar{h}_{v'}^c, \bar{h}_{v'}^b\}$. Then, we estimate the cosine distance $d_{vv'}^{cc}$ of $\bar{h}_v^c$ and $\bar{h}_{v'}^c$ for the shared class $c$, and the cosine distances $d_{vv'}^{ac}$ and $d_{vv'}^{cb}$ of the video features for $c$ and the non-shared classes $a$ and $b$.}
\label{fig:n-pair}
\end{figure}
This is formalized as minimizing the following n-pair loss in terms of three distances illustrated in Fig.~\ref{fig:n-pair}:%
\begin{align}\label{eq:ANPL}
    \mathcal{L}_{\text{NP}} =& \frac{1}{|C_{vv^{'}}|}\sum_{c \in C_{vv'}}\log\big[1 + \sum_{a \in C_v\setminus C_{vv'}}\exp(d_{vv'}^{cc}-d_{vv'}^{ac}) \nonumber \\
    &+\sum_{b \in C_{v'}\setminus C_{vv'}}\exp(d_{vv'}^{cc}-d_{vv'}^{cb})\big].
\end{align}
The distances in \eqref{eq:ANPL} are expressed in terms of the first-layer features  $\bm{h} \in \mathbb{R}^{n_h\times T}$, given by \eqref{eq:hidden}.  Specifically, features relevant for a particular class $c\in \mathcal{C}$, denoted as $\overline{\bm{h}}^c$, are computed by averaging $\bm{h}$ over all video frames labeled with $c$ in the MAP assignment. We consider both hard and soft class assignments to frames, and define:
\begin{eqnarray}
    \overline{\bm{h}}^c_{\text{hard}} &=& \frac{1}{\sum_{t=1}^{T}1(c=\hat{c}_t)}\sum_{t=1}^{T}\bm{h}[t]\cdot1(c=\hat{c}_t),  \label{eq:hard}\\
    \overline{\bm{h}}^c_{\text{soft}} &= &\sum_{t=1}^{T}\bm{h}[t]\cdot p(\bm{f}[c,t]),\label{eq:soft}
\end{eqnarray}
where $\hat{c}_t$ is the label predicted for $t$th frame by the SCV, and $p(\bm{f}[c,t])$ is the softmax score of $\bm{f}[c,t]$ specified in  \eqref{eq:secondlayer}.
Finally, we estimate the feature distances in \eqref{eq:ANPL} as
\begin{align}
    d_{vv'}^{cc'} = 1 - \frac{<\overline{\bm{h}}_v^{c},\overline{\bm{h}}_{v'}^{c'}>}{\|\overline{\bm{h}}_v^{c}\|
    \|\overline{\bm{h}}_{v'}^{c'}\|}.
\end{align}
where we omit the subscript ``hard'' or ``soft'', for simplicity.

Our total loss is a weighted average of $\mathcal{L}_{\text{CE}}$ and $\mathcal{L}_{\text{NP}}$:
\begin{align}
    \mathcal{L} = \lambda \mathcal{L}_{\text{CE}} + (1-\lambda)\mathcal{L}_{\text{NP}},
\end{align}
where we experimentally found that $\lambda=0.5$ is optimal.

\subsection{Complexity of Our Training}
Complexity of the SCV is $O(T^2|\mathcal{C}|^2)$, where $T$ is the video length, and $|\mathcal{C}|$ is the number of all action classes. The complexity mainly comes from the vanilla Viterbi algorithm in the first step of the SCV, whose complexity is $O(T^2|\mathcal{C}|^2)$. Complexity of the second step of the SCV is $O(T|\mathcal{C}|)$.

Complexity of the  n-pair loss is $O(T|\mathcal{C}|+|\mathcal{C}|^2)$, where $O(T|\mathcal{C}|)$ accounts for computation of the average feature $\overline{\bm{h}}_v^{c}$,  and $O(|\mathcal{C}|^2)$ accounts for estimation of the distances between features of shared and non-shared classes.

Therefore, our total training complexity is $O(T^2|\mathcal{C}|^2)$, whereas the training complexity of \cite{richard2018action} is $O(T|\mathcal{C}|)$.

\section{Inference on a Test Video}\label{sec:inference}

For inference on a test video, we follow the well-motivated Monte Carlo sampling of \cite{richard2018action}, which is not suitable for our training due to its prohibitive time complexity. As in \cite{richard2018action}, we make the assumption that the set of actions appearing in a test video has been seen in our set-supervised training, i.e., there is at least one training video with the same ground-truth set of actions as the test video. Our sampling has two differences as explained below.

Given a test video with length $T$, we begin by uniformly sampling action sets $C$ from all ground-truth sets of training videos. For every sampled $C$, we sequentially generate an action sequence $\bm{c}$ by uniformly sampling one action $c$ from $C$ at a time, allowing non-consecutive repetitions of actions along $\bm{c}$. The sampling stops when $\sum_{c\in\bm{c}}\lambda_c > T$. The first difference from \cite{richard2018action} is that we discard the obtained $\bm{c}$ if it does not include all actions from $C$, and proceed to randomly  generating a new $\bm{c}$. Thus, for every sampled $C$, we generate a number of legal  $\{\bm{c}\}$ that include all actions from $C$ and satisfy $\sum_{c\in\bm{c}}\lambda_c \le T$. The uniform sampling of sets and sequential generation of their respective legal action sequences is repeated until the total number of legal action sequences is  $K=\{\bm{c}\}=1000$.

 For every generated legal $\bm{c}$, as in \cite{richard2018action}, we use the standard dynamic programming to infer temporal extents $\bm{l}$ of actions from $\bm{c}$. The second difference is the evaluation of the new HMM's posterior as in \eqref{eq:Markov}, $p(\bm{c},\bm{l}|\bm{f})=\Big[\prod_{n=1}^{N-1} p({c_{n+1}}|c_n)\Big]  \Big[\prod_{n=1}^{N}p(l_n|c_n)\Big] \Big[\prod_{t=1}^{T}\frac{p(\bm{f}[c_{n(t)},t])}{p(c_{n(t)})}\Big]$. Out of $K$ candidate sequences, as the final solution, we select  $(\bm{c}^*,\bm{l}^*)$ that gives the maximum posterior $p(\bm{c}^*,\bm{l}^*|\bm{f})$.

 As in \cite{richard2018action}, complexity of our inference is  $O(T^2|\mathcal{C}|K)$, where $T$ is the video length, $|\mathcal{C}|$ is the number of action classes, $K$ is the number of candidate action sequences from the Monte Carlo sampling.


\section{Experiments}\label{sec:Experiments}

\subsection{Setup}
{\bf Datasets.} As in prior work \cite{richard2018action}, evaluation is done on the tasks of action segmentation and action alignment on three datasets, including Breakfast \cite{kuehne2014language},  MPII Cooking 2  \cite{rohrbach2016recognizing}, and Hollywood Extended (Ext)  \cite{bojanowski2014weakly}.

{\em Breakfast} consists of 1,712 videos with 3.6 million frames showing a total of 48 human actions for preparing breakfast. Each video has on average 6.9 action instances. For Breakfast, we report mean of frame accuracy (Mof) over the same 4-fold cross validation as in \cite{kuehne2014language}.

{\em MPII Cooking 2} consists of 273 videos with 2.8 million frames showing 67 human cooking actions.  We use the standard split of training and testing videos provided by \cite{rohrbach2012database}. For evaluation on MPII Cooking 2,  as in \cite{rohrbach2012database}, we use the midpoint hit criterion, \ie the midpoint of a correct detection segment should be within the ground-truth segment.

{\em Hollywood Ext} consists of 937 video clips with 800,000 frames showing 16 different classes. Each video has on average 2.5 action instances. We perform the same 10-fold cross validation as in \cite{bojanowski2014weakly}. For evaluation on Hollywood Ext, we use the Jaccard index, \ie intersection over detection (IoD), defined as $\text{IoD} = |GT\cap D|/|D|$, where $GT$ and $D$ denote the ground truth and detected action segments with the largest overlap.

{\bf Features.} For fair comparison, we use the same unsupervised video features as in \cite{richard2018action}. For all three datasets,  features are the framewise Fisher vectors of improved dense trajectories \cite{wang2013action}. The Fisher vectors for each frame are extracted over a sliding window of 20 frames. They are first projected to a 64-dimensional space by PCA, and then normalized along each dimension.

{\bf Training.} We train our model on a total number of 50,000 iterations. In each iteration, we randomly select two training videos that share at least one common action.  We set the initial learning rate as 0.01 and reduce it to 0.001 at the 10,000th iteration. Parameters of the dynamic HMM -- namely, dynamic transition probabilities, dynamic mean action length, and dynamic class prior per frame -- are initialized as for the static model, and then updated after each iteration as in \eqref{eq:transition dynamic}, \eqref{eq:dynamic length}, \eqref{eq:likelihood_dynamic}, respectively. In practice, it takes around 12 hours for training with one TITAN XP GPU.

{\bf Ablations.} We consider several variants of our approach for testing how its components affect performance:
\begin{itemize} [itemsep=-5pt,topsep=-5pt, partopsep=-1pt]
\item SCV = Our full approach with: the n-pair loss for regularization, the dynamic HMM whose parameters are given by \eqref{eq:transition dynamic}, \eqref{eq:dynamic length}, \eqref{eq:likelihood_dynamic}, and hard  class assignments to frames given by \eqref{eq:hard} for computing feature distances.
\item SCVnoreg = SCV, but without any regularization.
\item SCVbasereg = SCV, but with a simplified baseline regularization loss that minimizes only feature distances of shared classes by setting $d_{vv'}^{ac}=d_{vv'}^{cb}=0$ in \eqref{eq:ANPL}.
\item SCVsoft = SCV, but instead of hard we use soft class assignments to frames given by \eqref{eq:soft}.
\item SCVstatic = Our full approach as SCV, but instead of the dynamic HMM we use the static HMM whose parameters are given by \eqref{eq:transition static}, \eqref{eq:fixed length}, \eqref{eq:likelihood_static}.
\end{itemize}

\subsection{Action Segmentation}
This section presents our evaluation on action segmentation. Tab.~\ref{Table:action segmentation} shows our results on the three datasets. As can be seen, our SCV outperforms the state of the art by 6.9\% on Breakfast, 3.9\% on Cooking 2, and 8.4\% on Hollywood Ext, respectively. Tab.~\ref{Table:action segmentation} also shows results of the prior work with stronger supervision in training, where ground truth additionally specifies the temporal ordering of actions, called transcript. Surprisingly,  despite weaker supervision in training, our SCV outperforms some recent transcript-supervised approaches, e.g.,  ECTC \cite{huang2016connectionist} on Breakfast, and HMM+RNN \cite{richard2017weakly} and TCFPN \cite{ding2018weakly} on Hollywood Ext. The superior performance of SCV over the state of the art justifies our increase in training complexity.

\begin{figure}[t]
\begin{center}
\includegraphics[width=0.95\linewidth]{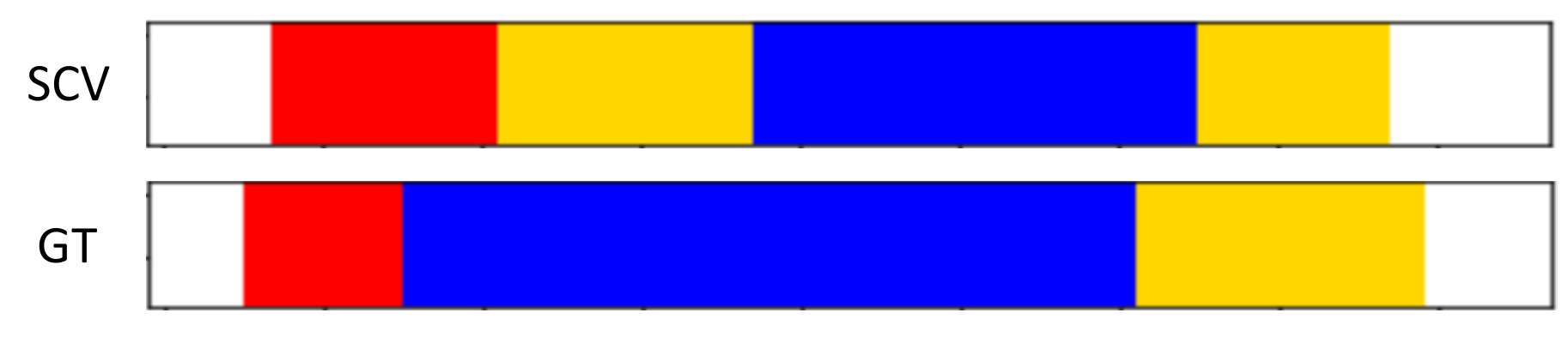}
\end{center}
   \caption{Action segmentation on an example test video \textit{P03\_cam01\_P03\_tea} from Breakfast. Top row: SCV result. Bottom row: ground truth with the color-coded action sequence \{\textcolor{red}{take\_cup}, \textcolor{blue}{add\_teabag}, \textcolor{mygold}{pour\_water}\}. The background frames are marked white. In general, SCV accurately detects true actions present in videos, but may miss their true ordering and locations.}
\label{fig:seg_example}
\end{figure}

Fig.~\ref{fig:seg_example} illustrates our action segmentation on an example test video \textit{P03\_cam01\_P03\_tea} from Breakfast.  As can be seen, SCV can detect true actions present in videos, but may miss their true ordering and true locations.

\begin{table}
\begin{center}
\begin{tabular}{l c c c}
\hline
  & Breakfast & Cooking2 & Holl.Ext \\
 Model  & (\textit{Mof}) & (\textit{midpoint hit}) & (\textit{IoD}) \\
\hline
\text{(Set-supervised)} \\
Action Set \cite{richard2018action} & 23.3 & 10.6 & 9.3\\
SCV& {\bf 30.2} & {\bf 14.5} & {\bf 17.7}\\
\hline
\multicolumn{2}{l}{\text{(Transcript-supervised)} }\\
OCDC \cite{bojanowski2014weakly} & 8.9 & - & - \\
HTK \cite{kuehne2017weakly} & 25.9 & 20.0 & 8.6\\
CTC \cite{huang2016connectionist} & 21.8 & - & -\\
ECTC \cite{huang2016connectionist} & 27.7 & - & -\\
HMM+RNN \cite{richard2017weakly} & 33.3 & - & 11.9\\
TCFPN \cite{ding2018weakly} & 38.4 & - & 18.3\\
NN-Viterbi \cite{richard2018neuralnetwork} & 43.0 & - & - \\
D3TW \cite{chang2019d3tw} & 45.7 & - & - \\
CDFL \cite{Li_2019_ICCV} & 50.2 & - & 25.8 \\
\hline
\end{tabular}
\end{center}
\caption{Weakly supervised action segmentation. Our SCV uses set-level supervision in training, while \cite{bojanowski2014weakly, kuehne2017weakly, huang2016connectionist, richard2017weakly, ding2018weakly, richard2018neuralnetwork} use stronger transcript-level supervision in training. The dash means that the prior work did not report the corresponding result. }
\label{Table:action segmentation}
\end{table}

{\bf Ablation --- Grammar.} As explained in Sec.~\ref{sec:inference}, we use Monte Carlo sampling to generate legal action sequences, called  Monte Carlo grammar. In Table.~\ref{Table:Grammar Comparison}, we test this component of our approach relative to the following baselines on Breakfast. The upper-bound baselines use a higher level of supervision --- transcripts of temporal orderings of actions made available for their inference. Thus, instead of generating the Monte Carlo grammar, the upper-bound baselines explicitly constrain their inference to search the solution only within the transcript grammar consisting of all ground-truth transcripts seen in their training. The lower-bound baselines do not consider legal action sequences in inference, but only predict actions from the provided set of all actions. Tab.~\ref{Table:Grammar Comparison} shows that SCV outperforms the state of the art for all types of grammars used in inference. From Tab.~\ref{Table:Grammar Comparison}, our Monte Carlo estimation of legal action sequences is reasonable, since the upper-bound SCV with the transcript grammar has an improvement of only 4.5\% on Breakfast.

\begin{table}
\begin{center}
\begin{tabular}{l c c c}
\hline
(Grammar)&\multicolumn{3}{c}{Breakfast (\textit{Mof})}\\
\cline{2-4}
Method & train & \qquad  \qquad &test \\
\hline
(No Grammar)  \qquad \qquad \\
Action Set \cite{richard2018action} & 14.7 &\qquad   \qquad &9.9\\
SCV & 16.6 &  &12.0\\
\hline
{(Monte Carlo Grammar)}  \qquad \qquad \\
Action Set \cite{richard2018action} & 28.2 &\qquad   \qquad  &23.3\\
SCV & 34.5 & & 30.2\\
\hline
{(Transcript Grammar)} \qquad \qquad \\
Action Set \cite{richard2018action} & 36.7 &\qquad  \qquad & 26.9\\
SCV & 40.5 &\qquad  \qquad & 34.7\\
\hline
\end{tabular}
\end{center}
\caption{Evaluation of our Monte Carlo sampling for generating legal action sequences (Monte Carlo Grammar) in inference relative to the upper-bound case that uses ground-truth transcripts of temporal orderings of actions (Transcript Grammar) in inference, and the lower-bound case that only predicts actions from the set of all actions in inference. Our SCV is superior for all grammars.}
\label{Table:Grammar Comparison}
\vskip 0.05in
\end{table}

\begin{table}
\begin{center}
\begin{tabular}{l c c c}
\hline
 & Breakfast & Cooking 2 & Holl.Ext \\
Model parameters & (\textit{Mof}) & (\textit{midpoint hit}) & (\textit{IoD}) \\
\hline
SCV+ground truth & 33.4 & 15.3 & 19.2 \\
\hline
SCVstatic & 28.7 & 13.7 & 16.1\\
SCV & 30.2 & 14.5 & 17.7\\
\hline
\end{tabular}
\end{center}
\caption{Evaluation of SCV when using ground-truth, static, and dynamic model parameters. SCV with ground-truth model parameters an upper-bound baseline.}
\label{Table:Comparison of length models.}
\vskip 0.05in
\end{table}

{\bf Ablation --- Static vs. Dynamic} Tab.~\ref{Table:Comparison of length models.} shows that our SCV with the dynamic HMM outperforms the variant with the static HMM -- SCVstatic. The table also shows the upper-bound performance of SCV when the mean action length $\lambda_c$ and class priors $p(c)$ are estimated directly from framewise ground-truth video labels. From Tab.~\ref{Table:Comparison of length models.}, our dynamic estimation of model parameters in \eqref{eq:transition dynamic}, \eqref{eq:dynamic length}, \eqref{eq:likelihood_dynamic} is very accurate, since SCV with the framewise ground truth model parameters has an improvement of only 3.2\% on Breakfast.

\begin{table}
\begin{center}
\begin{tabular}{l c c c}
\hline
 & Breakfast & Cooking 2 & Holl.Ext \\
Model  & (\textit{Mof}) & (\textit{midpoint hit}) & (\textit{IoD}) \\
\hline
SCVnoreg & 26.5 & 12.3 & 12.6\\
SCVbasereg & 27.8 & 13.1 & 14.8\\
SCVsoft & 29.3 & 14.2 & 16.1 \\
SCV & {\bf 30.2} & {\bf 14.5} & {\bf 17.7} \\
\hline
\end{tabular}
\end{center}
\caption{Evaluation of SCV with different regularizations. SCV gives the best results.}
\label{Table:Affinity loss Comparision.}
\end{table}

\begin{figure}[t]
\begin{center}
\includegraphics[width=0.95\linewidth]{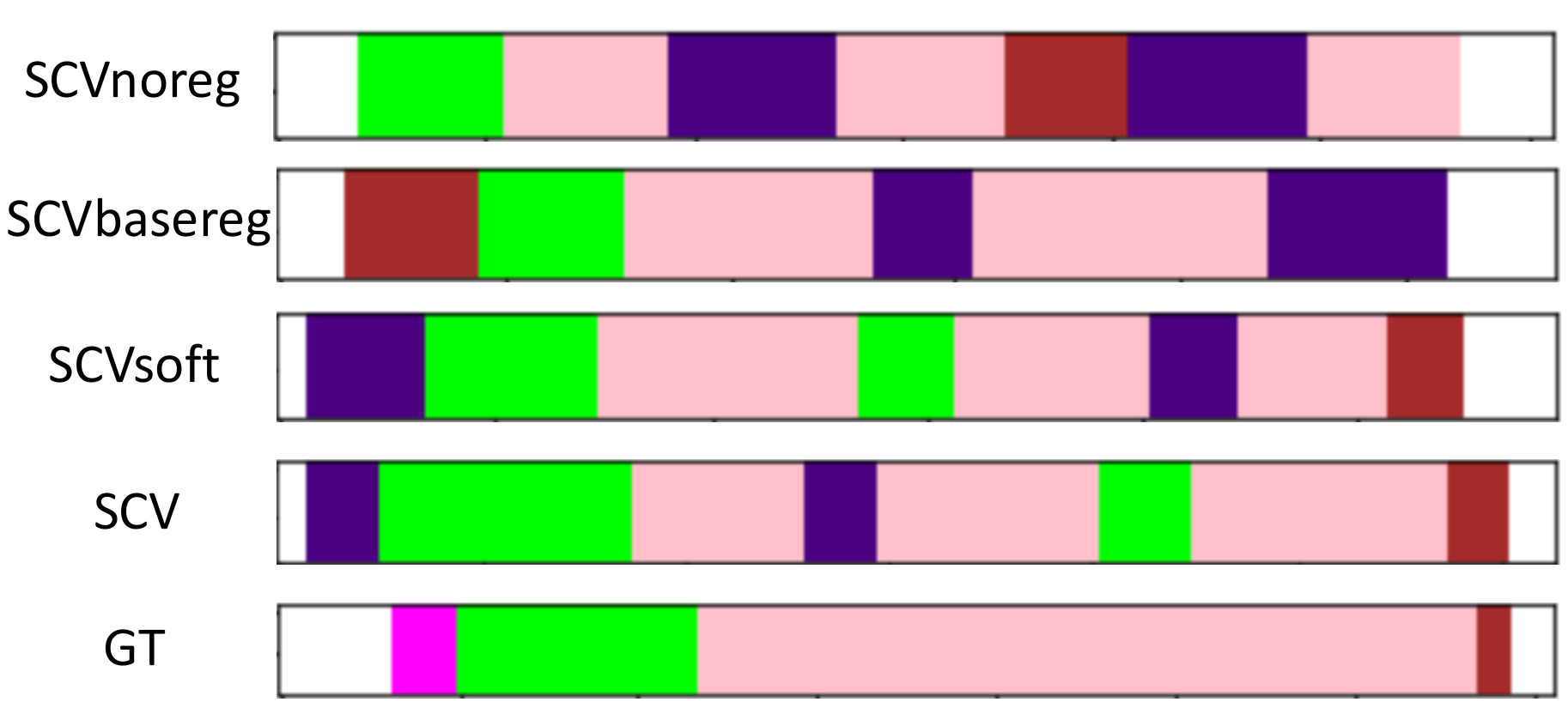}
\end{center}
   \caption{Our ablation of regularization on a sample test video \textit{P03\_webcam01\_P03\_friedegg} from Breakfast. The rows from top to bottom show action segmentation of: SCVnoreg, SCVbasereg, SCVsoft, SCV, and the ground-truth color-coded action sequence \{\textcolor{mymagenta}{crack\_egg},\textcolor{green}{fry\_egg},\textcolor{mypink}{take\_plate},\textcolor{mymaroon}{put\_egg2plate}\}. The background frames are marked white. SCV gives the best performance.}
\label{fig:npl_ablation}
\end{figure}

{\bf Ablation --- Regularization.} Tab.~\ref{Table:Affinity loss Comparision.} shows our evaluation of SCV when using different types of regularization, including no regularization.  As can be seen, SCV with the n-pair loss and the hard class assignments in training outperforms all of the other types of regularization. From Tab.~\ref{Table:Affinity loss Comparision.}, SCV outperforms SCVbasereg, which demonstrates that it is critical to maximize the discrimination margin between features of shared classes and features of non-shared classes, rather than only maximize affinity between features of shared classes. Fig.~\ref{fig:npl_ablation} illustrates our ablation of regularization on an example test video \textit{P03\_webcam01\_P03\_friedegg} from Breakfast. In general, SCV gives the best segmentation results.

\subsection{Action Alignment given Action Sets}
\begin{figure}[t]
\begin{center}
\includegraphics[width=0.95\linewidth]{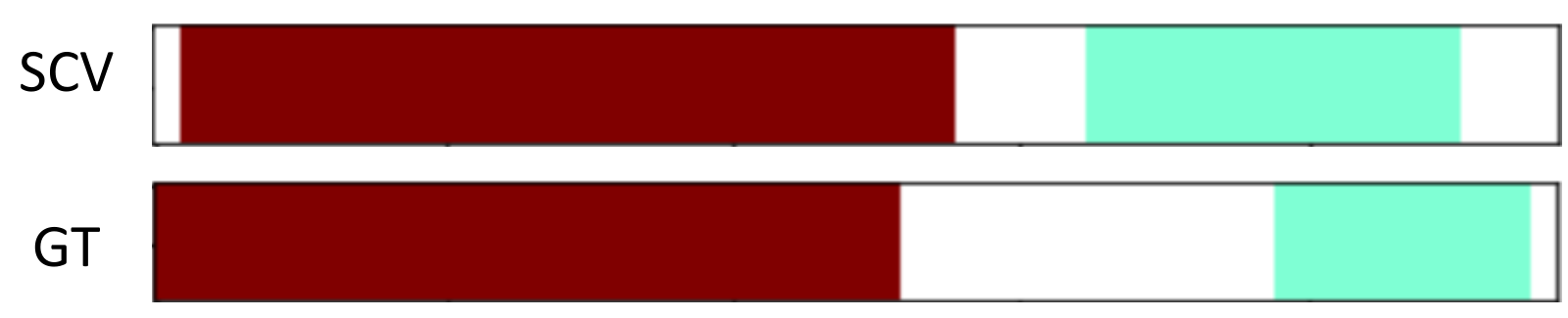}
\end{center}
   \caption{Action alignment on a sample test video \textit{1411} from Hollywood Ext. Top row: SCV result. Bottom row: ground truth with the color-coded action sequence \{\textcolor{mymaroon}{Run}, \textcolor{myaquamarine}{ThreatenPerson}\}.  The background frames are marked white. In general, SCV successfully aligns the actions, but may incorrectly detect their locations.}
\label{fig:align_example}
\end{figure}

The task of action alignment provides access to the ground-truth unordered set of actions $C$ present in a test video. Thus, instead of uniformly sampling sets of actions in inference on the test video, for alignment, we take the provided ground-truth $C$, and continue to perform Monte Carlo sampling of action sequences $\bm{c}$ that are restricted to come from $C$, as described in Sec.~\ref{sec:inference}. In comparison with the state of the art, Tab.~\ref{Table:alignment} shows that SCV improves action alignment by 12.4\% for Breakfast, 4.5\% for Cooking 2, 11.3\% for Hollywood Ext. Also, SCV achieves comparable results to some recent approaches that use the stronger transcript-level supervision in both training and testing. These results demonstrate that SCV is successful in estimating an optimal legal action sequence for a given ground-truth action set. This improved effectiveness of SCV over the state of the art justifies our reasonable increase in training complexity. Fig.~\ref{fig:align_example} illustrates action alignment on a sample test video \textit{1411} from Hollywood Ext. From the figure, SCV successfully aligns the actions, but may incorrectly detect their true locations.

\begin{table}
\begin{center}
\begin{tabular}{l c c c}
\hline
  & Breakfast & Cooking 2 & Holl.Ext. \\
  Model  & (\textit{Mof}) & (\textit{midpoint hit}) & (\textit{IoD}) \\
\hline
(Set-supervised) \\
Action Set \cite{richard2018action} & 28.4 & 10.6 & 24.2\\
SCV & {\bf 40.8}  & {\bf 15.1} & {\bf 35.5}\\
\hline
\multicolumn{2}{l}{\text{(Transcript-supervised)} }\\
ECTC \cite{huang2016connectionist} & $\sim$35 & - & $\sim$41\\
HTK \cite{kuehne2017weakly} & 43.9 & - & 42.4\\
OCDC \cite{bojanowski2014weakly} & - & - & 43.9\\
HMM+RNN \cite{richard2017weakly} & - & - & 46.3\\
TCFPN \cite{ding2018weakly} & 53.5 & - & 39.6\\
NN-Viterbi \cite{richard2018neuralnetwork} & - & - & 48.7 \\
D3TW \cite{chang2019d3tw} & 57.0 & - & 50.9 \\
CDFL \cite{Li_2019_ICCV} & 63.0 & - & 52.9 \\
\hline
\end{tabular}
\end{center}
\caption{Action alignment given ground-truth action sets of test videos. SCV outperforms the state of the art on the three datasets. The transcript-supervised prior work uses stronger supervision in training, yet SCV gives comparable results. The dash means that the prior work did not report the result.}
\label{Table:alignment}
\end{table}

\section{Conclusion}\label{sec:Conclusion}
We have addressed set-supervised action segmentation and alignment, and extended related work that trains their framewise classifiers with a pseudo ground truth generated by independently labeling video frames. We have made two contributions: (1) Set Constrained Viterbi algorithm aimed at generating a more accurate pseudo ground truth for our set-supervised training by efficiently approximating the NP-hard all-color shortest path problem; (2) Regularization of learning with the new n-pair loss aimed at maximizing a distance margin between video features of shared classes and video features of non-shared classes. Our evaluation demonstrates that we outperform the state of the art on three benchmark datasets for both action segmentation and alignment. This justifies an increase of our training complexity relative to that of prior work. Also, our approach gives comparable results to some related methods that use stronger transcript-level supervision in training. Our tests of various ablations and comparisons with reasonable baselines demonstrate effectiveness of  individual components of our approach.\\

\noindent{\bf{Acknowledgement}}. This work was supported by  DARPA  N66001-17-2-4029 and DARPA  N66001-19-2-4035 grants.

{\small
\bibliographystyle{ieee_fullname}
\bibliography{egbib}
}

\end{document}